\documentclass{svproc}

\usepackage{url}
\usepackage{graphicx}%
\usepackage{multirow}%
\usepackage{amsmath,amssymb,amsfonts}%
\usepackage{mathrsfs}%
\usepackage[title]{appendix}%
\usepackage{xcolor}%
\usepackage{textcomp}%
\usepackage{manyfoot}%
\usepackage{booktabs}%
\usepackage{algorithm}%
\usepackage{algorithmicx}%
\usepackage{algpseudocode}%
\usepackage{listings}%

\usepackage{subfig}
\usepackage[labelfont=bf]{caption}
\usepackage{rotating}
\usepackage{nicematrix}
\usepackage{makecell}
\usepackage{xcolor}
\usepackage{hyperref}
\usepackage[separate-uncertainty = true,multi-part-units=single]{siunitx}
\DeclareMathOperator*{\aggregate}{\text{AGGREGATE}}

\begin{document}
\mainmatter              
\title{Conditional Graph Neural Network for Predicting Soft Tissue Deformation and Forces}
\titlerunning{cGNN}  
%
\author{Madina Kojanazarova \and Florentin Bieder \and Robin Sandkühler \and Philippe C. Cattin}
\authorrunning{Madina Kojanazarova et al.} 
%
\tocauthor{Madina Kojanazarova, Florentin Bieder, Robin Sandkühler, Philippe C. Cattin}
\institute{University of Basel, Department of Biomedical Engineering, Allschwil, Switzerland\\
\email{M.Kojanazarova@unibas.ch}
}

\maketitle              

\begin{abstract}
Soft tissue simulation in virtual environments is becoming increasingly important for medical applications.
However, the high deformability of soft tissue poses significant challenges.
Existing methods rely on segmentation, meshing and estimation of stiffness properties of tissues.
In addition, the integration of haptic feedback requires precise force estimation to enable a more immersive experience.
We introduce a novel data-driven model, a \textit{conditional graph neural network} (cGNN) to tackle this complexity.
Our model takes surface points and the location of applied forces, and is specifically designed to predict the deformation of the points and the forces exerted on them.
We trained our model on experimentally collected surface tracking data of a soft tissue phantom and used transfer learning to overcome the data scarcity by initially training it with mass-spring simulations and fine-tuning it with the experimental data.
This approach improves the generalisation capability of the model and enables accurate predictions of tissue deformations and corresponding interaction forces.
The results demonstrate that the model can predict deformations with a distance error of \qty{0.35\pm0.03}{\mm} for deformations up to \qty{30}{\mm} and the force with an absolute error of \qty{0.37\pm0.05}{\N} for forces up to \qty{7.5}{\N}.
Our data-driven approach presents a promising solution to the intricate challenge of simulating soft tissues within virtual environments. Beyond its applicability in medical simulations, this approach holds the potential to benefit various fields where realistic soft tissue simulations are required.
\end{abstract}
\section{Introduction}
Virtual reality (VR) has significantly impacted medical applications, including pre-operative planning, intraoperative navigation, training, and patient rehabilitation \cite{Bin2020-dk,Preim2014-jb}. 
These immersive simulations enable medical professionals and trainees to interact with virtual anatomical structures, thereby improving understanding and increasing patient safety \cite{Dakson2017-qd,Pottle2019-lz}.
Combining haptic feedback systems with VR simulations enhances realism and interactivity, allowing users to manipulate virtual objects or tissues. Such realistic force feedback enables surgeons to perceive tissue resistance and other tactile sensations, leading to more effective training and improved surgical outcomes \cite{Coles2011}.
An example of such a visuo-haptic system was presented in \cite{Faludi2019-lf}, in which the patient-specific skull computed tomography (CT) volumes were visually and haptically rendered without resorting to mesh generation.

Accurately simulating soft tissue deformations is a significant challenge in VR-based anatomical simulations. 
Soft tissues exhibit complex mechanical behaviours, including elasticity, plasticity, and viscosity, making their simulation computationally demanding \cite{Holzapfel2001,Miller2000}.
Traditional biomechanical methods such as the finite element method (FEM) and mass-spring models (MSMs) have been employed to simulate soft tissue deformations \cite{Schill2001}. 
These methods often require time-consuming processes such as tissue segmentation from image volumes, mesh generation, and the estimation of tissue stiffness properties to simulate realistic behaviour \cite{Nguyen2020-tq,Payan2017-vi,Preim2014-jb}. 
Realistic haptic feedback simulations for deformable objects also rely on the outputs of the biomechanical methods \cite{Kim2003,Vaughan2014art}. 
Therefore, accurate and real-time interactive applications such as pre-operative planning are difficult to achieve. 
\subsubsection{Related Work}
Research on modelling soft tissue deformation has mainly focused on biomechanical methods such as FEM, MSMs, and their variants \cite{Nguyen2020-tq}. 
While FEM simulations can achieve high accuracies, they are computationally expensive and too slow for real-time visuo-haptic simulations \cite{Nguyen2020-tq}. 
In contrast, MSMs offer faster calculations suitable for real-time applications but are less accurate than FEM simulations because their parameters do not directly correspond to the biomechanical properties of soft tissues \cite{Nealen2006}. 
Several solutions have been proposed to adapt MSMs to the properties of soft tissues \cite{Golec2018-dd,Zerbato2007}. 

Recently, data-driven approaches have addressed different limitations, e.g. time limitations, of biomechanical models for estimating soft tissue characteristics. 
Early methods used regression models to learn the biomechanical behaviour of liver and breast tissue \cite{Martin-Guerrero2016-kl}, and to predict the deformation output of FEM simulations based on surface meshes of brain MR scans \cite{Tonutti2017-fu}. 
Other approaches have used image-based deep learning methods such as convolutional neural networks (CNNs) to predict deformations using FEM simulations of liver models for laparoscopic surgery simulation \cite{Mendizabal2020-gs,Mendizabal2020-jx,Pfeiffer2019-pn}. 

However, CNNs, which operate on regular grids, face challenges when applied to geometric data such as point clouds, graphs, and meshes, which are better suited for representing 3D objects with complex and irregular structures
\cite{Ahmed2018ASO,Bronstein2017-ec}.  
Graph neural networks (GNNs), in contrast, generalised convolutional operations on graph-like geometric data \cite{Scarselli2009-mt}, and current GNNs use the message passing architecture. 
While the applications of GNNs have mainly focused on classification, segmentation, and generation tasks on conventional graph-like structures, their application to soft tissue simulation is a novel idea. 
The authors of~\cite{Salehi2022-wy} proposed PhysGNN, which uses GraphSAGE \cite{Hamilton2017-ns} and GraphConv \cite{Morris2019-fb} operators to predict the deformation of brain tissue under applied forces. 
They simulated deformations with FEM and used meshes as input graphs along with force and specific tissue properties, demonstrating that GNNs can accurately predict soft tissue deformations with low computational time using GPU acceleration. 

\subsubsection{Proposed Method} 
The paper proposes a proof of concept for a data-driven method, namely a \textit{conditional Graph Neural Network} (cGNN), to simulate soft tissue deformations for VR applications with haptic feedback. 
Our method not only visually deforms the soft tissue at a given location but also estimates the interaction force, which is crucial for simulating haptic feedback in VR applications. 
By utilising a GNN, we leverage the power of point clouds and surface simulation to achieve accurate soft tissue modelling, overcoming the challenges associated with mesh generation, parameter estimation, and the preservation of detailed volumetric representations. 
To address the challenge of limited experimental data, we incorporated transfer learning. We initially pre-trained our model on simulated data generated using a mass-spring model (MSM) and fine-tuned it with experimentally collected surface tracking data of a soft tissue phantom. 
We aim to demonstrate that leveraging GNNs and transfer learning can make soft tissue simulation more accessible and efficient, especially with limited amounts of experimental data.

%
\section{Methods}\label{sec2}

\subsection{Experimental Data}
\begin{figure}
    \centering
    \subfloat[]{\includegraphics[width=0.30\textwidth]{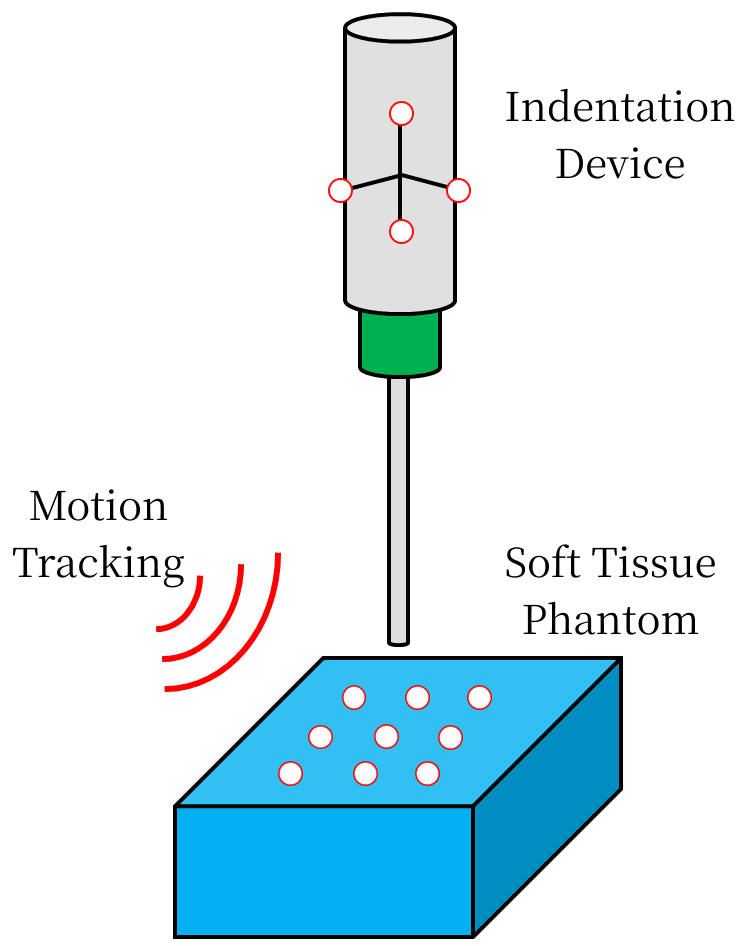}\label{setup}}
    \subfloat[]{\includegraphics[width=0.33\textwidth]{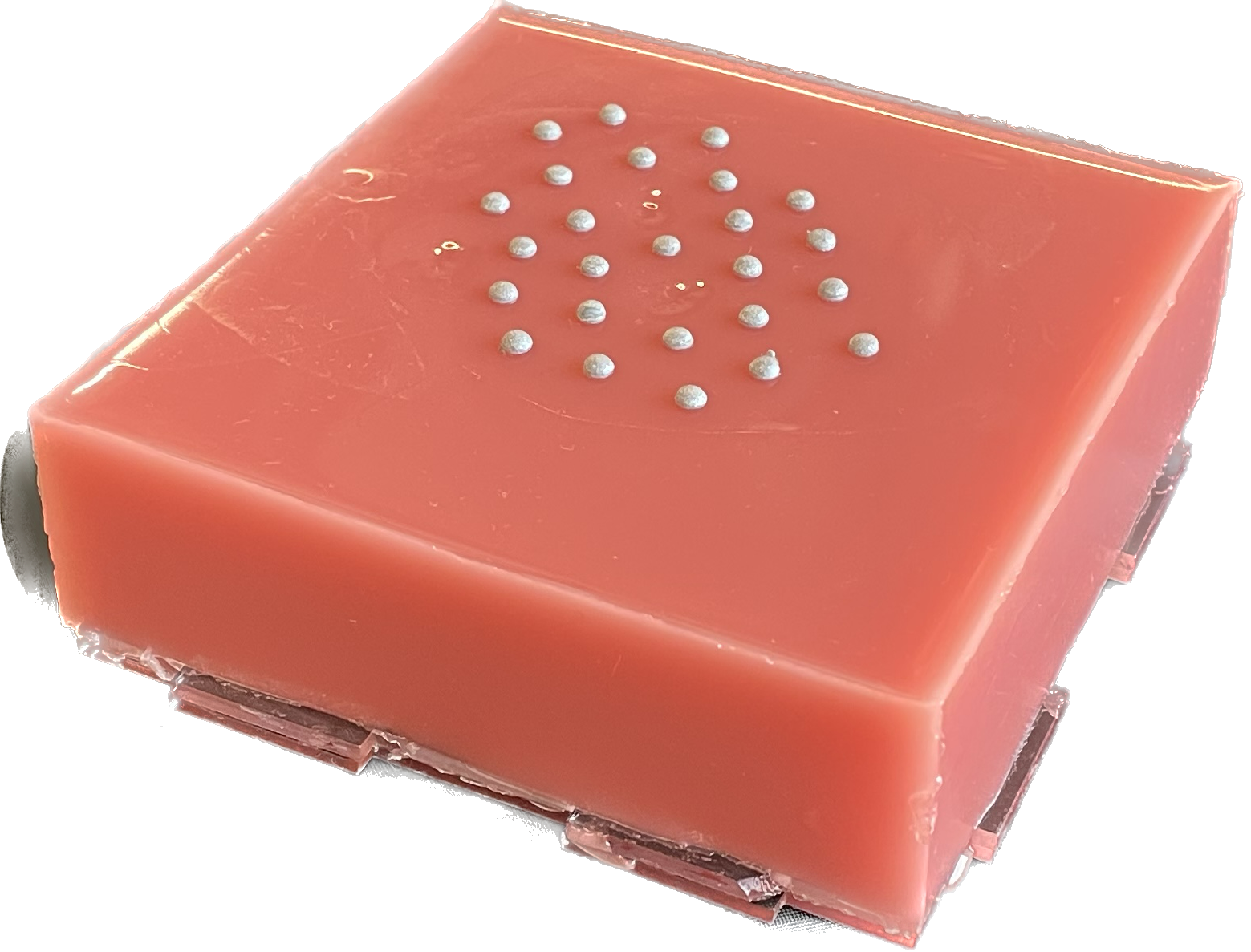}\label{silicone}}
    \subfloat[]{\includegraphics[width=0.3\textwidth]{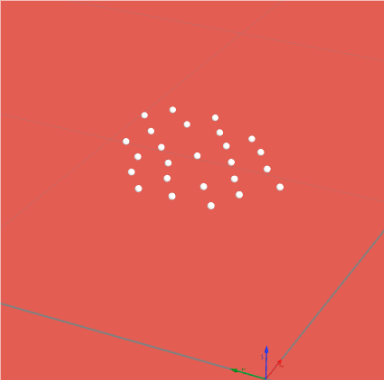}\label{tracked}}
    \caption{Experimental measurement. \textbf{a} System setup, \textbf{b} silicone phantom with markers, and \textbf{c} measured surface points} 
    \label{experimental}
\end{figure}

We acquired the geometric and force data using a silicone phantom with size \qtyproduct[product-units = power]{100 x 100 x 32}{\mm} by mixing the two components of silicone rubber (Eurosil 4 A+B, Schouten SynTec, the Netherlands) with an oil softener S (Eurosil Softener, Schouten SynTec, the Netherlands) in a 1A:1B:2.5S ratio \cite{Yushchenko2021}. 
The deformations were captured using $25$ pieces of \qty{3}{\mm} passive reflective markers (Fig.~\ref{silicone}) and a motion capture system consisting of five cameras (Qualisys AB, Sweden) at a sample rate of \qty{500}{\Hz} with a residual error of \qty{0.18\pm0.22}{\mm}. 
The force feedback was captured simultaneously using a Nano17 force sensor with the calibration SI-12-0.12 (ATI Industrial Automation, USA). 
The sensor was attached to an indentation device with four markers to track the movement (Fig.~\ref{setup}). 
The force measurement data was collected via a Speedgoat real-time target machine (Speedgoat, Switzerland) at a sample rate of \qty{1000}{\Hz}. 
The two measurements were synchronized by downsampling the force measurement to \qty{500}{\Hz}. 
We calculated the location of the tip of the indentation device using the Qualisys Track Manager 2021.2 and considered it the 26th marker on the surface.
We collected the experimental dataset four times at five indentation locations each and then extracted static deformations at every \qty{0.5}{\N} for training and testing.
This resulted in 244 static deformation states, with the max indentation force of \qty{6.2\pm1.4}{\N}. 

\subsection{Simulated MSM data}
\subsubsection{Mass-springs and parameter optimisation}
We simulated artificial soft tissue behaviour to create a larger dataset using mass-spring model (MSM) simulations. 
For this, we defined the soft tissue as a \qtyproduct[product-units = power]{100 x 100}{\mm} single-layer surface, simulated in a simple mass-spring model with linear between-mass springs and fixed-springs \cite{Chen2007-ls}. 
The surface consisted of $1024$ points organised in a triangular \numproduct{32 x 32} grid. 
With the masses set to \qty{0.00016}{\kg}, the damping factor $\num{0.1}$, and the simulation time step of \qty{0.0001}{\s}, we optimised the spring parameters to match the behaviour of the measured deformations by minimising the distance error.
The optimised parameters resulted in the between-mass spring coefficient \qty[per-mode = symbol]{100}{\newton\per\metre} and the fixed-spring coefficient \qty[per-mode = symbol]{21}{\newton\per\metre}.

\subsubsection{Data generation}
To deform the surfaces and acquire a range from small to large deformations, we applied forces of up to $F_{max}\qty{=7.5}{\newton}$ 
in $n_t\num{=15}$ time steps and 11 directions. The directions include the surface normal and 10 randomly sampled directions up to 45° from the surface normal. 
The forces were applied at 100 randomly chosen locations on the surface, each location affecting $n_n = 1$ mass. 
The affected surface point received the following amount of force at the time $t$:
\begin{equation} \label{eq:forces}
F_t = t \cdot \frac{1}{n_t \cdot n_n} F_{max},\quad \text{for} \quad t \in \{1, \ldots, n_t\}.
\end{equation}
We defined the system as stable when there were no significant changes in the velocity and the force of the masses $(v\qty[per-mode = symbol]{<0.02}{\metre\per\s},~ F\qty{<0.02}{\newton})$. The data generation resulted in \num{16500} static simulation examples. 

%
%

\subsection{Model}
We developed a \textit{conditional graph neural network} (cGNN) to predict the displacement of point clouds and corresponding forces.
The cGNN model utilises the DynamicEdgeConv graph-convolutional operator \cite{Wang2019-np} to capture the local geometric features of points while maintaining the permutation invariance.

Given a $D$-dimensional point cloud with $N$ points, DynamicEdgeConv constructs a directed graph $\mathcal{G} = (\mathcal{V}, \mathcal{E})$ of a local point cloud structure as a $k$-nearest neighbour ($k$-NN) graph with vertices $\mathcal{V}$ and edges $\mathcal{E}$. 
Each node in the graph also points to itself, making $\mathcal{G}$ a self-loop graph. 
Edge features in the graph are collected using an edge function $h$, a non-linear function with learnable parameters implemented as a multilayer perceptron (MLP). The messages from edge features are then aggregated with a symmetric aggregation operation (sum, mean, or max). 
The output of the DynamicEdgeConv at the point $x_i$ is therefore given by:
\begin{equation} \label{eq:mp}
x_i^\prime = \aggregate_{j \in \mathcal N(i)} (h(x_i,x_j)),
\end{equation}
where $\mathcal N(i) = \{j \in \mathcal V \mid  (i,j)\in \mathcal{E}\}$ is the neighbourhood of point $x_i$. 
\begin{figure}[b!]
    \includegraphics[width=\textwidth]{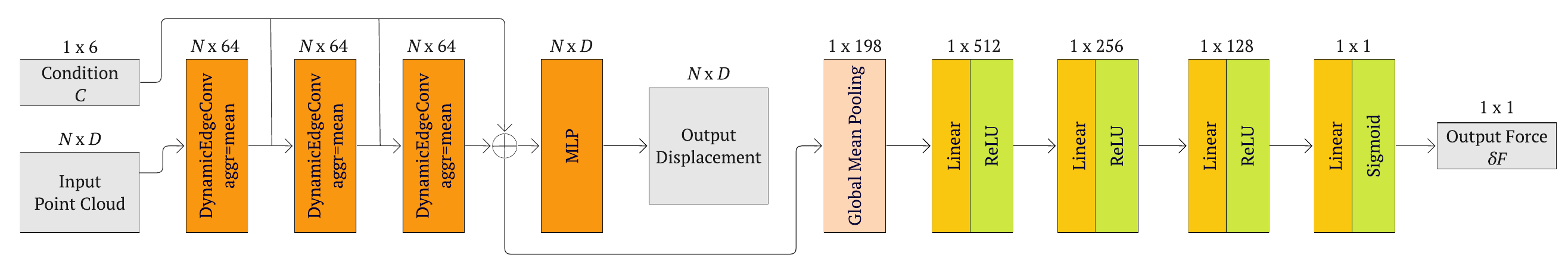}
    \caption{The architecture of the \textit{conditional graph neural network} (cGNN). $\oplus$ denotes the concatenation of the features} \label{architecture}
\end{figure}

Our model (Fig.~\ref{architecture}) takes a surface point cloud and extracts features through three DynamicEdgeConv operators. 
Concatenated outputs from DynamicEdgeConv layers and the specific conditions $C$ of deformation are then passed through an MLP, which learns to predict the Euclidean displacement field with the same size as the input. 
The predicted displacement field can be applied to the input surface to acquire a deformed surface. 
Furthermore, the network extracts the global feature information from previously learned features, followed by four linear layers and activation functions to predict the change in force magnitude applied to the surface. 

The input to the cGNN model are the $N$ surface points with each $D=3$ dimensions, given as Cartesian coordinates, and the deformation condition $C$.
The condition $C \in \mathbb R^{2D}$ includes the point tip coordinates of the start $c_s$ and end $c_e$ of the applied tissue deformation defined as 
$C = [c_s,~c_e]$.
For an input point cloud $x \in \mathbb R^{N \times D}$, the output of the model $f$ is given by $(\delta x,\delta F) = f(x,C)$,
where $\delta x \in \mathbb R^{N \times D}$ is the predicted displacement field and $\delta F \in \mathbb R$ is the predicted change in force magnitude.
The predicted displacement field is then applied to the input points to obtain the deformed surface $\hat{y} = x + \delta x$.
\subsection{Training Setup}
\subsubsection{Loss Function}
The learnable parameters of the model were trained by minimising the mean Euclidean distance $\mathcal{L}_d$ for deformed prediction, defined as:
\begin{equation} \label{eq:error}
\mathcal L_d = \frac1N\sum_{n=1}^N \Vert y_{n} - \hat{y}_{n} \Vert_2
\end{equation}
where $N$ denotes the number of points, $y_{n}$ is the position of the ground truth deformation, and $\hat{y}_{n}$ is the predicted deformation.
The Mean square error (MSE) was used as a loss function $\mathcal L_f$ for the force predictions. The total loss was $\mathcal L = \alpha \mathcal L_d + \mathcal L_f$.

\subsubsection{Hyperparameters}
We trained the model for $250$ epochs with a batch size of $32$, and used the Adam optimiser with an initial learning rate of $10^{-4}$.
The loss weight for $\mathcal L_d$ was set to $\alpha\num{=88}$. 
We used mean as the aggregation function in the DynamicEdgeConv layers in our model, and the number of neighbours $k$ for the computation of the $k$-NN-graph was set to $k\num{=5}$ as a trade-off between the model's prediction and time performance. 

\subsubsection{Infrastructure}
The MSM simulations were carried out on an AMD Epyc 7742 64-core CPU. The cGNN model was implemented using the open-source PyTorch-based machine learning library PyTorch Geometric 2.1.0. The models were trained and tested on an NVIDIA RTX 2080.
\subsubsection{Datasets and Transfer Learning}
We divided our datasets into two modes, \textit{Single-step} and \textit{Multi-step}. 
The Single-step set
includes all states of deformation from each time step $t$, with the ground truth as the deformation at $t\num[retain-explicit-plus]{+1}$ ($\delta F\qty{=0.5}{\newton}$).
The Multi-step set contains larger deformations and forces and includes 15 random states of $t$, where the ground truth was at least two time steps apart ($\delta F \qty{>0.5}{\newton}$).
Our model was trained on the combination of the two sets. 
These dataset modes were designed to train the model to handle various types of deformations and force increments.
We split the data on the level of the deformation locations into training:validation:test sets with a ratio of 7:2:1 for the MSM dataset, and with a ratio of 12:3:5 for the experimental dataset. 
In the experimental dataset, we performed a 5-fold cross-validation. 
Table~\ref{features} shows the test set features of the created datasets. 

We trained cGNN with the simulated MSM dataset and compared the results by testing on the experimental dataset with and without transfer learning (TL). 
In an ablation study, we tested the effect of the conditional encoding of our model by eliminating the deformation condition $C$ as an input, referred to as \textit{noCond}. 
To test the model's predictive consistency with respect to the number of training points used, we trained the model with augmented MSM data (aug MSM), by inputting \qty{10}{\%} of the randomly selected surface points for each sample during the training, and further testing it with the total amount of points. 

\begin{table}
\centering
\setlength{\tabcolsep}{0.5em}
\caption{Deformation depth features of the test sets, where $\delta F$ is the force range in the dataset and $\delta x$ is the displacement}\label{features}
\begin{tabular}{ll|ccc}
    \makecell[tl]{Dataset}  
    & \makecell[tl]{Mode} 
    & \makecell{$\delta F$ [N]} 
    & \makecell{$\delta x_{max}$ [mm]} 
    & \makecell{$\delta x_{mean}$ [mm]} \\
    \midrule
    \multirow{2}{*}{Experimental} & Single-step & $0.5$ & $12.91$ & $2.38\pm2.97$ \\ 
    & Multi-step & \numrange{1}{7.5} & $30.27$ & $8.91\pm7.96$ \\ 
    \midrule
    \multirow{2}{*}{MSM} & Single-step & $0.5$ & $6.76$ & $2.27\pm1.13$ \\ 
    & Multi-step & \numrange{1}{7.5} & $32.06$ & $11.93\pm6.66$ \\ 
\end{tabular}
\end{table}

\begin{figure}[b]
    \centering
    \includegraphics[width=0.99\textwidth]{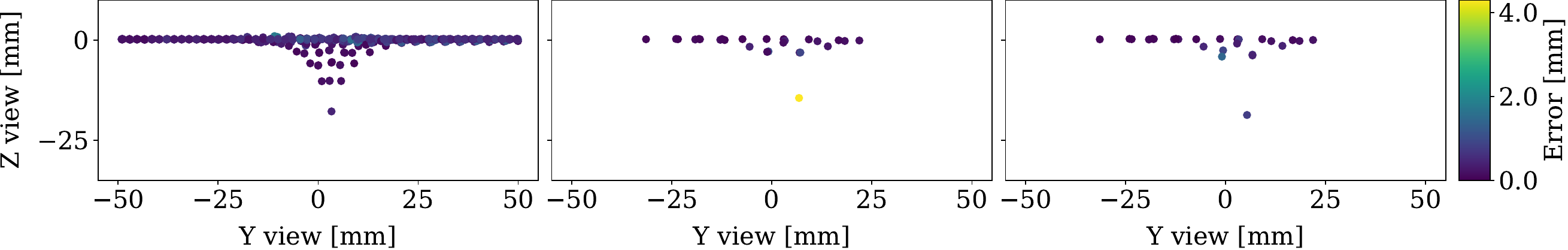}
    \captionsetup[subfigure]{justification=centering} 
    \captionsetup{labelformat=simple} 
    \captionsetup[subfigure]{labelformat=parens} 
    \captionsetup[subfigure]{position=top} 
    \captionsetup{labelsep=space} 
    \captionsetup{font=small} 
    \caption{Prediction results with ground truth force $\delta F\qty{=4.50}{\newton}$. Left: mass-spring simulation $\delta F_{pred}\qty{=4.49}{\newton}$. Middle: experimental data without transfer learning $\delta F_{pred}\qty{=4.65}{\newton}$. Right: experimental data with transfer learning $\delta F_{pred}\qty{=4.47}{\newton}$}
    \label{output_example}
\end{figure}

\section{Results and Discussion}
\subsubsection{Performance of cGNN}
The results in Table~\ref{tab_results} show that our model predicted the displacement with a \qty{0.43\pm0.11}{\mm} Euclidean error and an absolute force error of \qty{0.03\pm 0.06}{\newton} for the combined MSM dataset.
The training of our model without the conditional encoding (noCond) significantly decreased the performance for both displacement and force predictions, showing the importance of conditioning when generalising predictions. 
In contrast, training cGNN with the augmented MSM dataset (aug MSM), i.e. using \qty{10}{\%} of the randomly selected surface points for each sample, did not noticeably affect the performance when tested on the complete set of points. 

The implementation of TL for the experimental dataset significantly improved the deformation prediction compared to training without TL, resulting in a reduction of \qty{0.11}{\mm} (\qty{23}{\%}) of the mean displacement error and a reduction of \qty{1.03}{\mm} (\qty{34}{\%}) of the maximal displacement error with TL compared to without TL.
As for the force prediction, the performance was similar, resulting in lower force prediction errors with TL.
Figure~\ref{output_example} shows the deformation results where the predicted deformation field was applied to the input points.
Creating an MSM simulation took \qty{2.9321\pm0.5451}{\s}, while predicting it with the proposed cGNN took \qty{0.0947\pm0.0003}{\s}. 
For each experimental data prediction, the cGNN took only \qty{0.0029\pm0.0001}{\s}.

\begin{table}
\setlength{\tabcolsep}{0.3em}
\caption{Performance of cGNN on the mass-spring method dataset (MSM) and the experimental dataset (Exp) with and without transfer learning (TL). The model noCond indicates that cGNN was trained without conditional encoding}\label{tab_results}
\begin{tabular*}{\textwidth}{@{\extracolsep\fill}ll|cccc}
    \makecell{Model/\\ Train set}
    & \makecell{Test\\ set} 
    & \makecell{Force\\ MSE [$\text{N}^2$]} 
    & \makecell{Force\\Absolute\\Error [N]} 
    & \makecell{Mean\\ $\mathcal{L}_d$ [mm]}  
    & \makecell{Max\\ $\mathcal{L}_d$ [mm]} 
    \\
    \midrule
    noCond/MSM & MSM & $2.20$ & $1.19\pm0.88$ & $0.56\pm0.21$ & $5.43\pm5.64$ \\
    cGNN/MSM & MSM & \underline{$4\cdot10^{-3}$} & \underline{$0.03\pm0.06$} & \underline{$0.43\pm0.11$} & \underline{$2.22\pm2.11$} \\
    cGNN/aug MSM & MSM & $5\cdot10^{-3}$ & $0.03\pm0.06$ & $0.46\pm0.12$ & $2.35\pm2.16$ \\
    \midrule
    cGNN/Exp & Exp & $0.48$ & $0.48\pm0.08$ & $0.47\pm0.03$ & $3.06\pm0.55$ \\
    cGNN/MSM (TL) & Exp & $0.39$ & $0.41\pm0.02$ & $0.36\pm0.02$ & \textbf{2.03}$\pm$\textbf{0.29} \\
    cGNN/aug MSM (TL) & Exp & \textbf{0.35} & \textbf{0.37}$\pm$\textbf{0.05} & \textbf{0.35}$\pm$\textbf{0.03} & $2.10\pm0.35$ \\
\end{tabular*}
\end{table}

\subsubsection{Data Scarcity}
\begin{figure}[b]
    \centering
    \includegraphics[width=\textwidth]{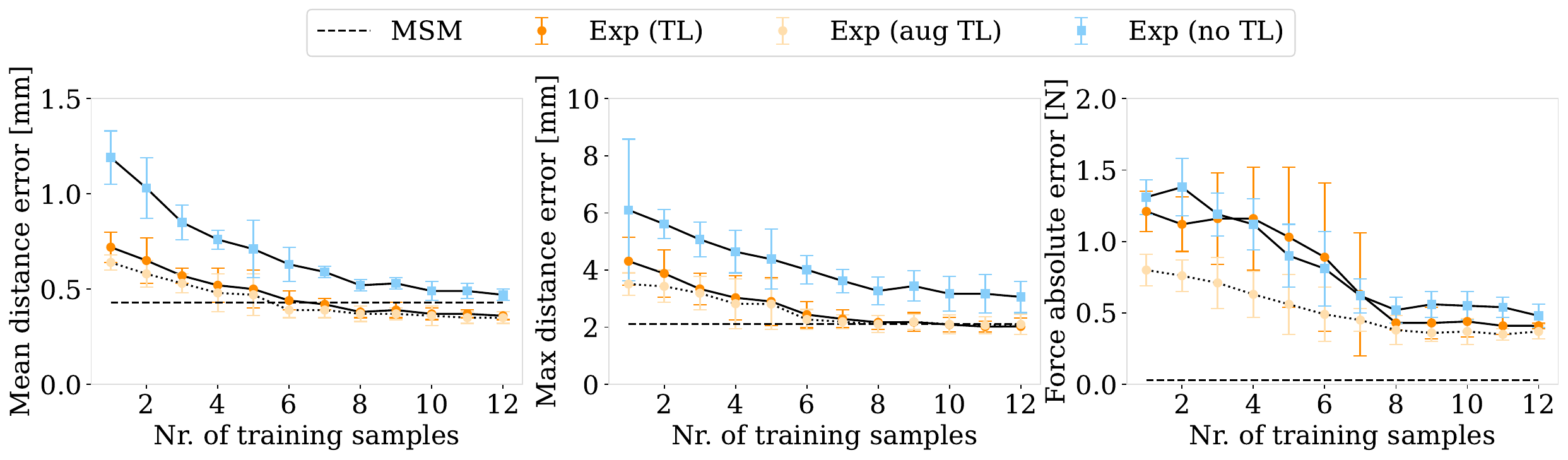}
    \caption{Mean distance error, max distance error and force absolute error of experimental data with transfer learning (Exp TL), without transfer learning (Exp no TL) and augmented transfer learning (Exp aug TL) when trained on different amounts of training samples compared to MSM simulations}
    \label{results}
\end{figure}

Due to the limited experimental data, we also trained cGNN with varying amounts of training samples, both with and without TL. 
Each training sample included all static data from one indentation location. 
As shown in Fig.~\ref{results}, using TL resulted in lower errors in deformation prediction, with optimal performance achieved using only 7-8 training samples. 
Additionally, TL with the model pre-trained on augmented MSM data demonstrated consistent performance for displacement predictions and significantly improved the force predictions. 



\subsubsection{Limitations and Implications}
The main limitation of this study is that we simplified our soft tissue model to a single surface layer of homogeneous tissues to avoid meshing. 
This simplification was chosen to demonstrate the feasibility of our approach and provide a clear proof of concept. 
Further analysis is needed to investigate the behaviour of inhomogeneous tissue with depth information, possibly using voxel intensities from CT volumes.

Additionally, the high computational time of our network when predicting MSM simulations with 1024 points is a significant limitation. 
This is primarily due to the DynamicEdgeConv layers, which involve expensive calculations of dynamic edge features in each layer. 
Despite this, we chose this graph-convolutional layer to develop a model capable of simulating deformations based solely on surface points, thereby eliminating the need for meshing the object. 
Notably, while the computation time for the simulated data was high, it was reasonable for the experimental data due to the smaller number of points. 
This reflects a realistic scenario where experimental data often involves fewer points, indicating that computation time is not a critical issue in practical applications.

The use of transfer learning on limited experimental data highlights an important implication of our work. 
We demonstrated that our model, initially trained on simulated data, could be effectively fine-tuned with a small set of experimental data. 
This approach mitigates the issue of data scarcity, showing that even with limited data, the model can be adapted to new datasets. 
Our ``data scarcity'' test revealed that the model performs well with as few as 7-8 examples, suggesting that future applications can easily adopt this approach when additional data becomes available.

It is important to note that our work and results cannot be directly compared with other studies due to the different approaches and data used. 
Other methods, such as \cite{Salehi2022-wy}, used larger meshes and FEM simulations, where computation time was not an issue, but they required the simulation of biomechanical properties. 
Their model requires an input of tissue properties and the applied force to predict deformations. 
In contrast, our method predicts deformations and forces without requiring detailed tissue information, simplifying the process and avoiding the computational complexity associated with biomechanical property estimation.
Our study demonstrates a versatile approach to soft tissue surface deformation that can predict the force exerted on the tissue without requiring additional information about the tissue properties. 

\section{Conclusions}
This study aimed to propose a novel graph-based and data-driven simulation method, cGNN, to generalise soft tissue deformation for VR applications that use haptic force feedback. 
We demonstrated the ability of cGNN to successfully deform synthetic and real soft tissue surfaces using a transfer learning approach. 
Our model also predicts the force applied to them by conditioning them to the start and end location of the exerted force. 
By showing that our method can accurately predict deformations using a simple single-layer model, we provide a foundation for future developments in more complex tissue simulations. 
This proof of concept highlights the potential of our approach to enhance the realism and effectiveness of VR-based medical training and planning tools.

\subsubsection{Acknowledgments}
The authors would like to acknowledge the support from Felix Erb and the BIROMED-Lab (Department of Biomedical Engineering, University of Basel) for the assistance with the experimental data measurement. 

%
%
%



\end{document}